\documentclass[11pt,a4paper]{article}
\usepackage{emnlp2018}
\usepackage{times}
\usepackage{latexsym}

\usepackage{url}

\usepackage{graphicx}
\usepackage{tikz}
\usepackage{multicol}
\usepackage{amsmath}
\usepackage{amssymb}
\usepackage{pgfplots}
\usepackage{subcaption}
\usepackage{standalone}
\usepackage[utf8]{inputenc}
\usepackage{multicol}
\pgfplotsset{compat=1.14}

\usepackage{resizegather}

\usetikzlibrary{shapes}
\def\verticalstep{1}

\aclfinalcopy 
\def\aclpaperid{***} 

\DeclareMathOperator{\attn}{\mathcal{A}}

\DeclareMathOperator{\concat}{concat}
\DeclareMathOperator*{\softmax}{softmax}

\newcommand{\veryshortarrow}[1][3pt]{\mathrel{%
   \vcenter{\hbox{\rule[-.5\fontdimen8\textfont3]{#1}{\fontdimen8\textfont3}}}%
   \mkern-4mu\hbox{\usefont{U}{lasy}{m}{n}\symbol{41}}}}

\title{Input Combination Strategies for Multi-Source Transformer Decoder}

\author{Jindřich Libovický \and Jindřich Helcl \and David Mareček \\
  Charles University, Faculty of Mathematics and Physics \\
  Institute of Formal and Applied Linguistics \\
  Malostransk\' e n\' am\v est\' i 25, 118 00 Prague, Czech Republic \\
  {\tt \{libovicky, helcl, marecek\}@ufal.mff.cuni.cz}}

\date{}

\begin{document}
\maketitle

\begin{abstract}
In multi-source sequence-to-sequence tasks, the attention mechanism can be
modeled in several ways.
This topic has been thoroughly studied on recurrent architectures.
In this paper, we extend the previous work to the encoder-decoder attention in
the Transformer architecture.
We propose four different input combination strategies for the encoder-decoder
attention: serial, parallel, flat, and hierarchical.
We evaluate our methods on tasks of multimodal translation and translation with
multiple source languages.
The experiments show that the models are able to use multiple sources and
improve over single source baselines.
\end{abstract}

\section{Introduction}


The Transformer model \citep{vaswani2017attention} recently demonstrated
superior performance in neural machine translation (NMT) and other
sequence generation tasks such as text summarization or image captioning
\citep{kaiser2017one}. However, all of these setups consider only a single
input to the decoder part of the model.

In the Transformer architecture, the representation of the source
sequence is supplied to the decoder through the encoder-decoder attention.
This attention sub-layer is applied between the self-attention and
feed-forward sub-layers in each Transformer layer. Such arrangement
leaves many options for the incorporation of multiple encoders.

So far, attention in sequence-to-sequence learning with multiple source
sequences was mostly studied in the context of recurrent neural networks (RNNs).
\citet{libovicky2017attention} explicitly capture the distribution over
multiple inputs by projecting the input representations to a shared vector space and
either computing the attention over all hidden states at once, or hierarchically, using
another level of attention applied on the context vectors.
\citet{zoph2016multi} employ a gating mechanism for combining the context vectors.
\citet{voita2018context} adapted the gating mechanism for use within the Transformer
model for context-aware MT.
The other aproaches are however not directly usable in the Transformer model.

We propose a number of strategies of combining the different sources in the
Transformer model. Some of the strategies described in this work are an
adaptation of the strategies previously used with recurrent neural networks
\citep{libovicky2017attention}, whereas the rest of them is a novel
contribution devised for the Transformer architecture. We test these strategies
on multimodal machine translation (MMT) and multi-source machine translation
(MSMT) tasks.

This paper is organized as follows. In Section ~\ref{sec:transformer},
we briefly describe the decoder part of the Transformer model.
We propose a number of input combination strategies for the multi-source
Transformer model in Section ~\ref{sec:strategies}.
Section ~\ref{sec:experiments} describes the experiments we performed, and
Section ~\ref{sec:results} shows the results of quantitative evaluation.
An overview of the related work is given in Section ~\ref{sec:related}.
We discuss the results and conclude in Section ~\ref{sec:conclusions}.

\section{Transformer Decoder}
\label{sec:transformer}



The Transformer architecture is based on the use of attention. Attention, as
conceptualized by \citet{vaswani2017attention}, can be viewed as a soft-lookup
function operating on an associative memory. For each query vector in query set $Q$, the attention
computes a set of weighted sums of values $V$ associated with a set of keys $K$,
based on their similarity to the query.

The variant of the attention function used in the Transformer architecture is
called \emph{multi-head scaled dot-product} attention.
%
%
Scaled dot-product of queries and keys is used as the similarity measure.
Given the dimension of the input vectors $d$, the attention is computed as follows:
\begin{gather}
%
%
\attn(Q, K, V) = \softmax \left( \dfrac{Q K^\top}{\sqrt{d}}\right)V.
\label{eq:singlehead}
\end{gather}
%
%
In the multi-head variant, the vectors that represent the queries, keys,
and values are linearly transformed to a number of projections (usually
with smaller dimension), called \emph{attention heads}.
The attention is computed in each head independently and the outputs are
concatenated and projected back to the original dimension:
\begin{gather}
\attn^h(Q,K,V) = \sum_{i=1}^h C_i W^O_i
\label{eq:context}
\end{gather}
where $W^O_i \in \mathbb{R}^{d_h \times d}$ are trainable parameter matrices
used as projections of the attention head outputs of dimension $d_h$ to the
model dimension $d$, and
\begin{gather}
C_i = \attn(QW^Q_i, KW^K_i, VW^V_i)
\end{gather}
where $W^Q$, $W^K$, and $W^V \in \mathbb{R}^{d \times d_h}$, are trainable
projection matrices used to project the attention inputs to the attention heads.

The model itself consists of a number of layers, each of which is divided in
three sub-layers: self-attention, encoder-decoder (or cross) attention,
and a feed-forward layer. Both of the attention types use identical sets
for keys and values. The states of the previous layer are used as the query set.
The self-attention sub-layer attends to the previous
decoder layer (i.e. the sets of queries and keys are identical). Since the decoder works
autoregressively from left to right, during training, the self-attention is masked
to prevent attending to the future positions in the sequence.
The encoder-decoder attention sub-layer attends to the final layer of the encoder.
The feed-forward sub-layer consists of a single non-linear projection (usually to a space with
larger dimension), followed by a linear projection back to the vector space with the
original dimension.
The input of each sub-layer is summed with the output,
creating a residual connection chain throughout the whole layer stack.




\section{Proposed Strategies}\label{sec:strategies}

\begin{figure}

\begin{subfigure}[t]{\linewidth}
\centering
\hspace*{-1.5cm}
\def\verticalstep{1}

\begin{tikzpicture}[]

\draw[fill,outer sep=0,inner sep=0]  (0,0) circle (1pt) node (before_self) {};
\draw (0,-0.5) -- (before_self);

\draw (0,1 * \verticalstep) node[rectangle,draw] (self_att) {self-attention};
\draw[inner sep=0]  (0,2 * \verticalstep) node (after_self) {$\oplus$};

\draw[->] (before_self) -- (self_att);
\draw[->] (before_self) .. controls ++(2.5,0.4 * \verticalstep) and ++(2.5,0) .. (after_self);
\draw[->] (self_att) -- (after_self);

\draw (0,3 * \verticalstep) node[rectangle,draw] (cross_att_1) {cross-attention \# 1};
\draw (-4,3 * \verticalstep) node[shape=ellipse,draw,style=dashed] (encoder_1) {encoder \# 1};
\draw[inner sep=0]  (0,4 * \verticalstep) node (after_cross_1) {$\oplus$};

\draw[->] (encoder_1) to[out=340,in=220] (cross_att_1);
\draw[->] (after_self) -- (cross_att_1);
\draw[->] (cross_att_1) -- (after_cross_1);
\draw[->] (after_self) .. controls ++(3.5,0.1 * \verticalstep) and ++(3.5,0) .. (after_cross_1);

\draw (0,5  * \verticalstep) node[rectangle,draw] (cross_att_2) {cross-attention \# 2};
\draw (-4,5  * \verticalstep) node[shape=ellipse,draw,style=dashed] (encoder_2) {encoder \# 2};
\draw[inner sep=0]  (0,6  * \verticalstep) node (after_cross_2) {$\oplus$};

\draw[->] (encoder_2) to[out=340,in=220] (cross_att_2);
\draw[->] (after_cross_1)-- (cross_att_2);
\draw[->] (cross_att_2) -- (after_cross_2);
\draw[->] (after_cross_1) .. controls ++(3.5,0.1 * \verticalstep) and ++(3.5,0) .. (after_cross_2);

\draw (0,7  * \verticalstep) node[rectangle,draw] (ff) {feed-forward layer};
\draw[inner sep=0]  (0,8  * \verticalstep) node (after_ff) {$\oplus$};

\draw[->] (after_cross_2) -- (ff);
\draw[->] (after_cross_2) .. controls ++(3.5,0.1 * \verticalstep) and ++(3.5,0) .. (after_ff);
\draw[->] (ff) -- (after_ff);

\draw[->] (after_ff) -- ++(0,0.5);

\end{tikzpicture}
\caption{serial}\label{fig:serial}
\end{subfigure}
\vspace{10pt}

\begin{subfigure}[t]{\linewidth}
\centering
\hspace*{0.6cm}
\def\verticalstep{1}

\begin{tikzpicture}[]

\draw[fill,outer sep=0,inner sep=0]  (0,0) circle (1pt) node (before_self) {};
\draw (0,-0.5) -- (before_self);

\draw (0,1 * \verticalstep) node[rectangle,draw] (self_att) {self-attention};
\draw[inner sep=0]  (0,2 * \verticalstep) node (after_self) {$\oplus$};

\draw[->] (before_self) -- (self_att);
\draw[->] (before_self) .. controls ++(2.3,0.4 * \verticalstep) and ++(2.1,0) .. (after_self);
\draw[->] (self_att) -- (after_self);

\draw (-2,3 * \verticalstep) node[rectangle,draw] (cross_att_1) {cross-attention \# 1};
\draw (-3.5,1.7 * \verticalstep) node[shape=ellipse,draw,style=dashed] (encoder_1) {encoder \# 1};
\draw[inner sep=0]  (0,4 * \verticalstep) node (after_cross_1) {$\oplus$};

\draw[->] (encoder_1) to[out=90,in=220] (cross_att_1);
\draw[->] (after_self) -- (cross_att_1);
\draw[->] (cross_att_1) -- (after_cross_1);
\draw[->] (after_self) .. controls ++(6.5,0.4 * \verticalstep) and ++(6.5,0) .. (after_cross_1);

\draw (2,3  * \verticalstep) node[rectangle,draw] (cross_att_2) {cross-attention \# 2};
\draw (3.5,1.7  * \verticalstep) node[shape=ellipse,draw,style=dashed] (encoder_2) {encoder \# 2};

\draw[->] (encoder_2) to[out=90,in=320] (cross_att_2);
\draw[->] (after_self) -- (cross_att_2);
\draw[->] (cross_att_2) -- (after_cross_1);

\draw (0,5  * \verticalstep) node[rectangle,draw] (ff) {feed-forward layer};
\draw[inner sep=0]  (0,6  * \verticalstep) node (after_ff) {$\oplus$};

\draw[->] (after_cross_1) -- (ff);
\draw[->] (after_cross_1) .. controls ++(3.5,0.1 * \verticalstep) and ++(3.5,0) .. (after_ff);
\draw[->] (ff) -- (after_ff);

\draw[->] (after_ff) -- ++(0,0.5);

\end{tikzpicture}
\caption{parallel}\label{fig:parallel}
\end{subfigure}
\vspace{10pt}

\begin{subfigure}[t]{\linewidth}
\centering
\hspace*{-1.5cm}
\def\verticalstep{1}

\begin{tikzpicture}[]

\draw[fill,outer sep=0,inner sep=0]  (0,0) circle (1pt) node (before_self) {};
\draw (0,-0.5) -- (before_self);

\draw (0,1 * \verticalstep) node[rectangle,draw] (self_att) {self-attention};
\draw[inner sep=0]  (0,2 * \verticalstep) node (after_self) {$\oplus$};

\draw[->] (before_self) -- (self_att);
\draw[->] (before_self) .. controls ++(2.5,0.4 * \verticalstep) and ++(2.5,0) .. (after_self);
\draw[->] (self_att) -- (after_self);

\draw (-4,3 * \verticalstep) node[shape=ellipse,draw,style=dashed] (encoder_1) {encoder \# 1};
\draw (-4,1  * \verticalstep) node[shape=ellipse,draw,style=dashed] (encoder_2) {encoder \# 2};
\draw (-2,2  * \verticalstep) node[shape=rectangle,draw,style=dashed] (concat) {concat};
\draw[->] (encoder_1) to[out=290,in=180] (concat);
\draw[->] (encoder_2) to[out=70,in=180] (concat);

\draw (0,3 * \verticalstep) node[rectangle,draw] (cross_att_1) {cross-attention};
\draw[inner sep=0]  (0,4 * \verticalstep) node (after_cross_1) {$\oplus$};

\draw[->] (concat) to[out=0,in=220] (cross_att_1);
\draw[->] (after_self) -- (cross_att_1);
\draw[->] (cross_att_1) -- (after_cross_1);
\draw[->] (after_self) .. controls ++(3.5,0.1 * \verticalstep) and ++(3.5,0) .. (after_cross_1);

\draw (0,5  * \verticalstep) node[rectangle,draw] (ff) {feed-forward layer};
\draw[inner sep=0]  (0,6  * \verticalstep) node (after_ff) {$\oplus$};

\draw[->] (after_cross_1) -- (ff);
\draw[->] (after_cross_1) .. controls ++(3.5,0.1 * \verticalstep) and ++(3.5,0) .. (after_ff);
\draw[->] (ff) -- (after_ff);

\draw[->] (after_ff) -- ++(0,0.5);

\end{tikzpicture}
\caption{flat}\label{fig:flat}
\end{subfigure}
\vspace{10pt}

\begin{subfigure}[t]{\linewidth}
\centering
\hspace*{0.6cm}
\def\verticalstep{1}

\begin{tikzpicture}[]

\draw[fill,outer sep=0,inner sep=0]  (0,0) circle (1pt) node (before_self) {};
\draw (0,-0.5) -- (before_self);

\draw (0,1 * \verticalstep) node[rectangle,draw] (self_att) {self-attention};
\draw[inner sep=0]  (0,2 * \verticalstep) node (after_self) {$\oplus$};

\draw[->] (before_self) -- (self_att);
\draw[->] (before_self) .. controls ++(2.3,0.4 * \verticalstep) and ++(2.1,0) .. (after_self);
\draw[->] (self_att) -- (after_self);

\draw (-2,3 * \verticalstep) node[rectangle,draw] (cross_att_1) {cross-attention \# 1};
\draw (-3,1.7 * \verticalstep) node[shape=ellipse,draw,style=dashed] (encoder_1) {encoder \# 1};
\draw[->] (encoder_1) to[out=90,in=220] (cross_att_1);
\draw[->] (after_self) -- (cross_att_1);

\draw (2,3  * \verticalstep) node[rectangle,draw] (cross_att_2) {cross-attention \# 2};
\draw (3.3,1.7  * \verticalstep) node[shape=ellipse,draw,style=dashed] (encoder_2) {encoder \# 2};
\draw[->] (encoder_2) to[out=90,in=320] (cross_att_2);
\draw[->] (after_self) -- (cross_att_2);

\draw (0,4.5  * \verticalstep) node[rectangle,draw] (hier) {attention over contexts};
\draw[->] (cross_att_1) -- (hier);
\draw[->] (cross_att_2) -- (hier);

\draw[inner sep=0]  (0,5.5 * \verticalstep) node (after_cross) {$\oplus$};
\draw[->] (after_self) .. controls ++(6,0.4 * \verticalstep) and ++(6,0) .. (after_cross);
\draw[->] (hier) -- (after_cross);

\draw (0,6.5  * \verticalstep) node[rectangle,draw] (ff) {feed-forward layer};
\draw[inner sep=0]  (0,7.5  * \verticalstep) node (after_ff) {$\oplus$};

\draw[->] (after_cross) -- (ff);
\draw[->] (after_cross) .. controls ++(3,0.1 * \verticalstep) and ++(3,0) .. (after_ff);
\draw[->] (ff) -- (after_ff);

\draw[->] (after_ff) -- ++(0,0.5);

\end{tikzpicture}
\caption{hierarchical}\label{fig:hier}
\end{subfigure}

\caption{Schemes of computational steps for the serial, parallel, flat, and
hierarchical attention combination in a single layer of the decoder.}
\label{fig:architecture}

\end{figure}
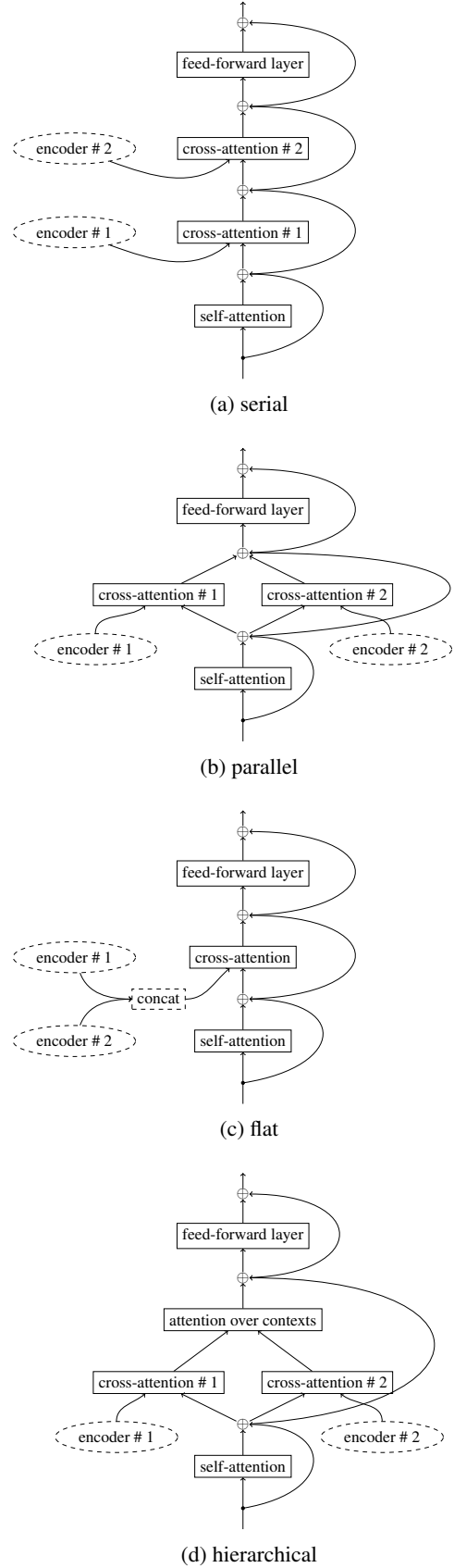


We propose four input combination strategies for multi-source variant of the Transformer
network, as illustrated in Figure~\ref{fig:architecture}. Two of them, serial and
parallel, model the encoder-decoder attentions independently and are a
natural extension of the sub-layer scheme in the transformer decoder. The other
two versions, flat and hierarchical, are inspired by approaches proposed for RNNs by
\citet{libovicky2017attention} and model joint distributions over the inputs.


\paragraph{Serial.}
The serial strategy (Figure~\ref{fig:serial}) computes the
encoder-decoder attention one by one for each input encoder.
The query set of the first cross-attention is the set of the context vectors
computed by the preceding self-attention. The query set of each subsequent
cross-attention is the output of the preceding sub-layer. All of these
sub-layers are interconnected with residual connections.


\paragraph{Parallel.}
In the parallel combination strategy (Figure~\ref{fig:parallel}), the model
attends to each encoder independently and then sums up the context vectors.
Each encoder is attended using the same set of queries, i.e. the output of the self-attention
sub-layer. Residual connection link is used between the queries and the summed
context vectors from the parallel attention.
\begin{gather}
%
%
\attn^h_{\mathit{para}}(Q, K_{1:n}, V_{1:n}) = \sum_{i=1}^{n} \attn^h(Q, K_i, V_i)
\end{gather}

\paragraph{Flat.}
The encoder-decoder attention in the flat combination strategy (Figure~\ref{fig:flat})
uses all the states of all input encoders as a single set of keys and values.
Thus, the attention models a joint distribution over a flattened set of all encoder
states. Unlike the approach taken in the recurrent setup \citep{libovicky2017attention},
where the flat combination strategy requires an explicit projection of
the encoder states to a shared vector space, in the Transformer models,
the vector spaces of all layers are tied with residual connections. Therefore, the
intermediate projection of the states of each encoder is not necessary.
\begin{gather}
%
K_{\mathit{flat}} = V_{\mathit{flat}} = \concat_i(K_i)
\\
\attn^h_{\mathit{flat}}(Q, K_{1:n}, V_{1:n}) = \attn^h(Q, K_\mathit{flat}, V_\mathit{flat})
\end{gather}

\paragraph{Hierarchical.}
In the hierarchical combination (Figure~\ref{fig:hier}), we first compute the attention
independently over each input. The resulting contexts are then treated as
states of another input and the attention is computed once again over these
states.
\begin{gather}
%
%
K_{\mathit{hier}} = V_{\mathit{hier}} = \concat_i( \attn^h(Q, K_i, V_i) )
\\
\attn^h_{\mathit{hier}}(Q, K_{1:n}, V_{1:n}) = \attn^h(Q, K_\mathit{hier}, V_\mathit{hier})
\end{gather}

\section{Experiments}
\label{sec:experiments}



We conduct our experiments on two different tasks: multimodal translation and
multi-source machine translation. We use Neural Monkey
\citep{neuralMonkey}\footnote{\url{http://github.com/ufal/neuralmonkey}} for
design, training, and evaluation of the experiments.

In all experiments, the encoder part of the network follows the Transformer architecture as
described by \citet{vaswani2017attention}.

We optimize the model parameters using Adam optimizer \citep{kingma2014adam}
with initial learning rate 0.2, and Noam learning rate decay \citep{vaswani2017attention}
with $\beta_1$ = 0.9,  $\beta_2 = 0.98$, $\epsilon = 10^{-9}$, and 4,000 warm-up steps.
The size of a mini-batch size of 32 for MMT, and 24 for multi-source
MT experiments.

During decoding, we use beam search of width 10 and length normalization of
1.0 \citep{wu2016google}.

\subsection{Multimodal Translation}

The goal of MMT \citep{specia2016shared} is translating
image captions from one language into another given both the source and image
as the input. We use Multi30k dataset \citep{elliot2016multi} containing
triplets of images, English captions and their English translations into
German, French and Czech. The dataset contains 29k triplets for training, 1,014
for validation and a test set of 1,000. We experiment with all language pairs
available in this dataset.

We extract image feature using the last convolutional layer of the ResNet
network \citep{he2016deep} trained for ImageNet classification. We apply a linear
projection into 512 dimensions on the image representation, so it has the same
dimension as the rest of the model. For each language pair, we create a shared
wordpiece-based vocabulary of approximately 40k subwords. We share the embedding matrices
across the languages and we use the transposed embedding matrix as the output projection
matrix as proposed by \citet{press2017tieembeddings}.

We use 6 layers in the textual encoder and decoder, and set the model dimension to 512. We set the
dimension of the hidden layers in the feed-forward sub-layers to 4096. We use
16 heads in the attention layers.

During the evaluation, we follow the preprocessing used in WMT Multimodal Translation
Shared Task \citep{specia2016shared}.

Conclusions of previous work show \citep{elliott2017imagination} that the improved
performance of the multimodal models
compared to textual models can come from improving the input representation.
In order to test whether it is also the case with our models or the models explicitly use
the visual input, we perform an adversarial
evaluation similar to \citet{elliott2018adversairal}.
We evaluate the model while providinng a random image and observe
how it affects the score and observe whether their quality drops.

\subsection{Multi-Source MT}

In this set of experiment, we attempt to generate a sentence in a target
language, given equivalent sentences in multiple source languages.

We use the Europarl corpus \citep{tiedemann2012opus} for training and testing
the MSMT\@. We use Spanish, French, German, and English as source
languages and Czech as a target language. We selected an intersection of the
bilingual sub-corpora using English as a pivot language. Our dataset contains
511k 5-tuples of sentences for training, 1k for validation and another 1k for
testing.

Due of the memory demands of having four encoders, we use
a smaller model than in the previous experiment. The encoders only have 4
layers and the decoder has 6 layers with embeddings size 256,
feed-forward layers dimension 2048, and 8 attention heads. We
use a shared word-piece vocabulary of 48k subwords. As in the MMT experiments,
the transposition of the embedding matrix is reused as the parameters of the output
projection layer \citep{press2017tieembeddings}.

We use bilingual English-to-Czech translation as a single source baseline. The
baseline uses vocabulary of 42k subwords from Czech and English only.

Similarly to the MMT, we also perform adversarial evaluation.
To evaluate the importance of the source languages for the translation quality,
when randomizing one of the source languages.

\section{Results}\label{sec:results}

\begin{table*}[ht]
\newcommand{\R}[2]{#1 \tiny $\pm$ \small #2}
\begin{center}
\scalebox{0.87}{%
\begin{tabular}{l|ccc|ccc|ccc}
 & \multicolumn{3}{c|}{MMT: en$\veryshortarrow$de}
 & \multicolumn{3}{c|}{MMT: en$\veryshortarrow$fr}
 & \multicolumn{3}{c}{MMT: en$\veryshortarrow$cs} \\
 & B\scalebox{.8}{LEU} & M\scalebox{.8}{ETEOR} & \scalebox{.8}{adv.}B\scalebox{.8}{LEU}
 & B\scalebox{.8}{LEU} & M\scalebox{.8}{ETEOR} & \scalebox{.8}{adv.}B\scalebox{.8}{LEU}
 & B\scalebox{.8}{LEU} & M\scalebox{.8}{ETEOR} & \scalebox{.8}{adv.}B\scalebox{.8}{LEU}
\\ \hline

 baseline
 & \R{38.3}{.8} & \R{56.7}{.7} & --- 
 & \R{59.6}{.9} & \R{72.7}{.7} & --- 
 & \R{30.9}{.8} & \R{29.5}{.4} & --- 
 \\ \hline
serial
 & \R{38.7}{.9} & \R{57.2}{.6} & \R{37.3}{.6} 
 & \R{60.8}{.9} & \R{75.1}{.6} & \R{58.9}{.9} 
 & \R{31.0}{.8} & \R{29.9}{.4} & \R{29.7}{.8} 
 \\
parallel
 & \R{38.6}{.9} & \R{57.4}{.7} & \R{38.2}{.8} 
 & \R{60.2}{.9} & \R{74.9}{.6} & \R{58.9}{.9} 
 & \R{31.1}{.9} & \R{30.0}{.4} & \R{30.4}{.8} 
\\
flat
 & \R{37.1}{.8} & \R{56.5}{.6} & \R{35.7}{.8} 
 & \R{58.0}{.9} & \R{73.3}{.7} & \R{57.0}{.9} 
 & \R{29.9}{.8} & \R{29.0}{.4} & \R{28.2}{.8} 
\\
hierarchical
 & \R{38.5}{.8} & \R{56.5}{.6} & \R{38.1}{.8} 
 & \R{60.8}{.9} & \R{75.1}{.6} & \R{60.2}{.9} 
 & \R{31.3}{.9} & \R{30.0}{.4} & \R{31.0}{.8} 
\\
\end{tabular}}
\end{center}

\caption{Quantitative results of the MMT experiments on the 2016 test set.
Column `adv. BLEU' is an adversarial evaluation with randomized image input.}
\label{tab:mmt_results}

\end{table*}

We evaluate the results using BLEU \citep{papineni2002bleu} and
METEOR \citep{denkowski2011meteor} as implemented in MultEval.
\footnote{\url{https://github.com/jhclark/multeval}} The results of the MMT
task are tabulated in Table~\ref{tab:mmt_results}. The results of the multi-source MT
are shown in Table~\ref{tab:msmt_results}.

In MMT, the input combination significantly surpassed the text-only baseline
in English-to-French translation.
The performance in other target languages is only slightly better than the textual
baseline.

The only worse score was achieved by the flat combination strategy. We hypothesize
this might be because the optimization failed to find a common representation of the
input modalities that could be used to compute the joint distribution.

The adversarial evaluation with randomly selected input images shows that all our models rely on
both inputs while generating the target sentence and that providing incorrect visual input harms the
model performance. The modality gating in the hierarchical
attention combination seems to make the models more robust to noisy visual input.

In the multi-source translation task, all the proposed strategies perform better than
single-source translation from English to Czech. Among the combination strategies, the
best-scoring is the serial stacking of the attentions.
In multimodal translation, the flat combination has shown to be the
best-performing strategy.

\begin{figure}[t!]
\begin{center}
\includegraphics[width=\linewidth]{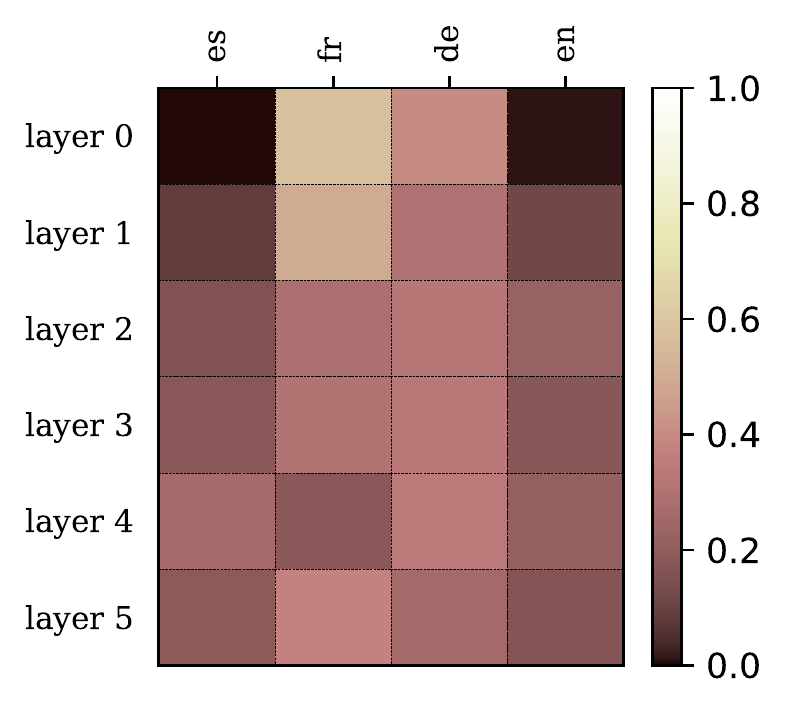}
\end{center}
\caption{Attention over contexts in the hiearchical strategy over the decoder layers.}
\label{fig:hier-att-over-contexts}
\end{figure}

Analysis of the attention distribution
shows that the serial strategy use information from all source languages. The
parallel strategy almost does not use the Spanish source and the flat strategy
prefers the English source. The hierarchical strategy uses information from all
source languages, however the attentions are sometimes more fuzzy than in the
previous strategies. Figure~\ref{fig:hier-att-over-contexts} shows what source languages
were attended on different layers of the encoder. Other examples of the attention
visualization are shown in Appendix~\ref{sec:supplemental}.

The adversarial evaluation shows all the models used English as a primary source.
Providing incorrect English source harms. Introducing noise into other languages
affects the score in much smaller scale.

\begin{table*}[ht]
\newcommand{\R}[2]{#1 \tiny $\pm$ \small #2}
\begin{center}
\scalebox{0.87}{%
\begin{tabular}{l|cc|cccc}
 & \multicolumn{2}{c|}{MSMT}
& \multicolumn{4}{c}{Adversarial evaluation (BLEU)}\\
 & B\scalebox{.8}{LEU} & M\scalebox{.8}{ETEOR} &
   en & de & fr & es \\ \hline

 baseline
 & \R{16.5}{.5} & \R{20.5}{.3} 
 & ---
 & ---
 & ---
 & ---
 \\ \hline
serial
 & \R{20.5}{.6} & \R{23.5}{.5} 
 & \R{ 8.1}{.4}                 
 & \R{19.7}{.5}                 
 & \R{19.5}{.6}                 
 & \R{18.4}{.5}                 
 \\
parallel
 & \R{20.5}{.6} & \R{23.3}{.3} 
 & \R{ 1.4}{.2}                 
 & \R{18.7}{.5}                 
 & \R{17.9}{.5}                 
 & \R{20.3}{.5}                 
\\
flat
 & \R{20.4}{.6} & \R{23.3}{.3}  
 & \R{ 0.2}{.1}
 & \R{19.9}{.6}
 & \R{20.0}{.6}
 & \R{19.6}{.5}
\\
hierarchical
 & \R{19.4}{.5} & \R{22.7}{.3}
 & \R{ 4.2}{.3}
 & \R{18.3}{.5}
 & \R{18.3}{.5}
 & \R{15.3}{.5}
\\
\end{tabular}}
\end{center}

\caption{Quantitative results of the MMT experiment. The adversarial evaluation
shows the BLEU score when one input language was changed randomly.}
\label{tab:msmt_results}

\end{table*}

\section{Related Work}\label{sec:related}

MMT was so far solved only within the RNN-based architectures.
\citet{elliott2015multi} report significant improvements with a non-attentive
model. With attentive models \citep{bahdanau2014neural}, the additional visual
information usually did not improve the models significantly
\citep{caglayan2016multimodality,helcl2017cuni} in terms of BLEU score.
Our models slightly outperform these models in the single model setup.

Except for using the image features direct input to the model, they can be used
as an auxiliary objective \citep{elliott2017imagination}. In this setup, the
visually grounded representation, improves the MMT significantly, achieving
similar results that our models achieved using only the Multi30k dataset.

To our knowledge, multi-source MT has also been studied only using the RNN-based
models. \citet{dabre2017enabling} use simple concatenation of source sentences
in various languages and process them with a single multilingual encoder.

\citet{zoph2016multi} try context concatenation and hierarchical gating method for
combining context vectors in attention models with multiple inputs encoded by separate encoders.
In all of their experiments, the multi-source methods significantly surpass the
single-source baseline. \citet{nishimura2018multi} extend the former approach
for situations when of the source languages is missing, so that the
translation system does not overly rely on a single source language like
some of the models presented in this work.

\section{Conclusions}\label{sec:conclusions}

We proposed several input combination strategies for multi-source
sequence-to-sequence learning using the Transformer model \citep{vaswani2017attention}. Two of the strategies
are a straightforward extension of cross-attention in the Transformer model:
the cross-attentions are combined either serially interleaved by residual
connections or in parallel. The two remaining strategies are an adaptation of the flat and the hierarchical attention combination strategies introduced by \citet{libovicky2017attention} in context of recurrent sequence-to-sequence models.

The results on the MMT task show similar properties an in
RNN-based models \citep{caglayan2017lium,libovicky2017attention}.
Adding visual features significantly improves translation into French and brings
minor improvements on other language pairs.
All the attention combinations perform similarly with the exception of the flat strategy
which probably struggles with learning a shared representation of the input tokens
and the image representation.

Evaluation on multi-source MT shows significant improvements
over the single-source baseline. However, the adversarial evaluation suggests that
the model relies heavily on the English input and only uses the additional source languages
for minor modifications of the output. All attention combinations performed similarly.

\section*{Acknowledgments}
This research received support from
the grant No. 18-02196S and No. P103/12/G084 of the Grant Agency of the Czech Republic,
and the grant No. 976518 of the Grant Agency of the Charles University.
This research was partially supported by SVV project number 260~453.

\bibliography{emnlp2018}
\bibliographystyle{acl_natbib_nourl}

\onecolumn

\vspace{-0.6cm}\begin{multicols}{2}
\appendix
\section{Attention Visualizations}
\label{sec:supplemental}
We show cross-attention visualizations for the four proposed combination strategies
on Multi-source MT. The Czech target wordpieces are in rows, the source Spanish,
French, German, and English wordpieces are concatenated and shown in columns.
These attentions were taken form the decoder's fourth layer and were averaged
across the individual heads. For serial and parallel strategy the cross-attention
weights sum to one for each language separately, the flat strategy has only one common
cross-attention, and for the hierarchical strategy visualization the cross-attention
weights for individual languages are multiplied by the weights
of the attention over contexts.
\end{multicols}
\begin{center}
\includegraphics[width=0.92\textwidth]{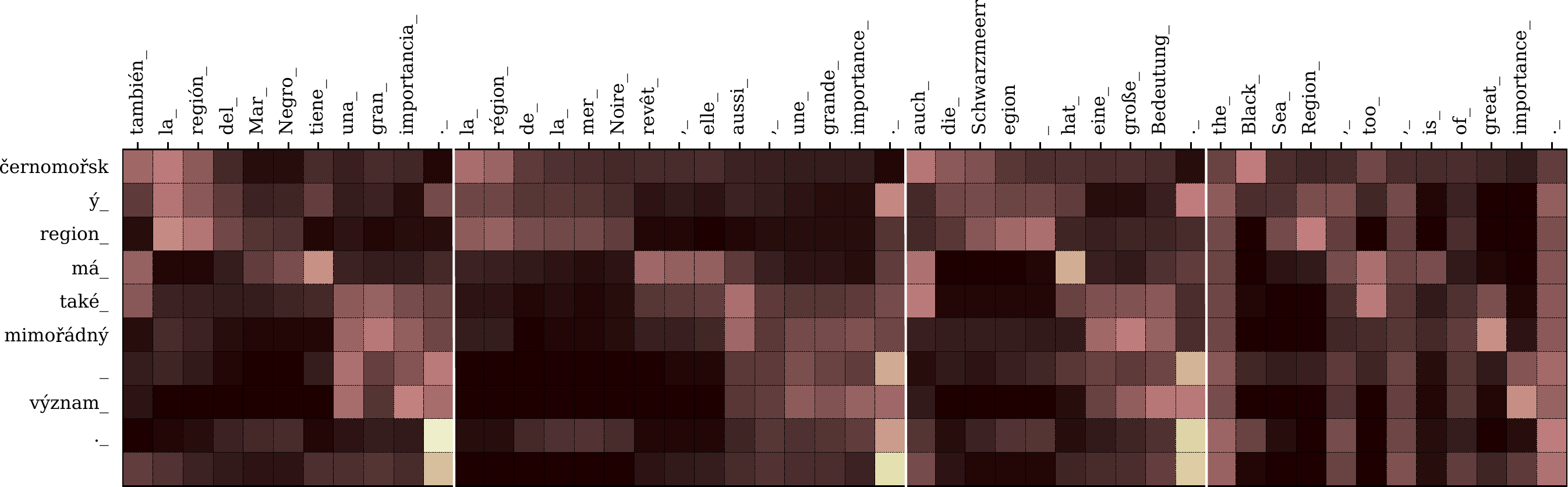} \\
a) serial \\
\includegraphics[width=0.92\textwidth]{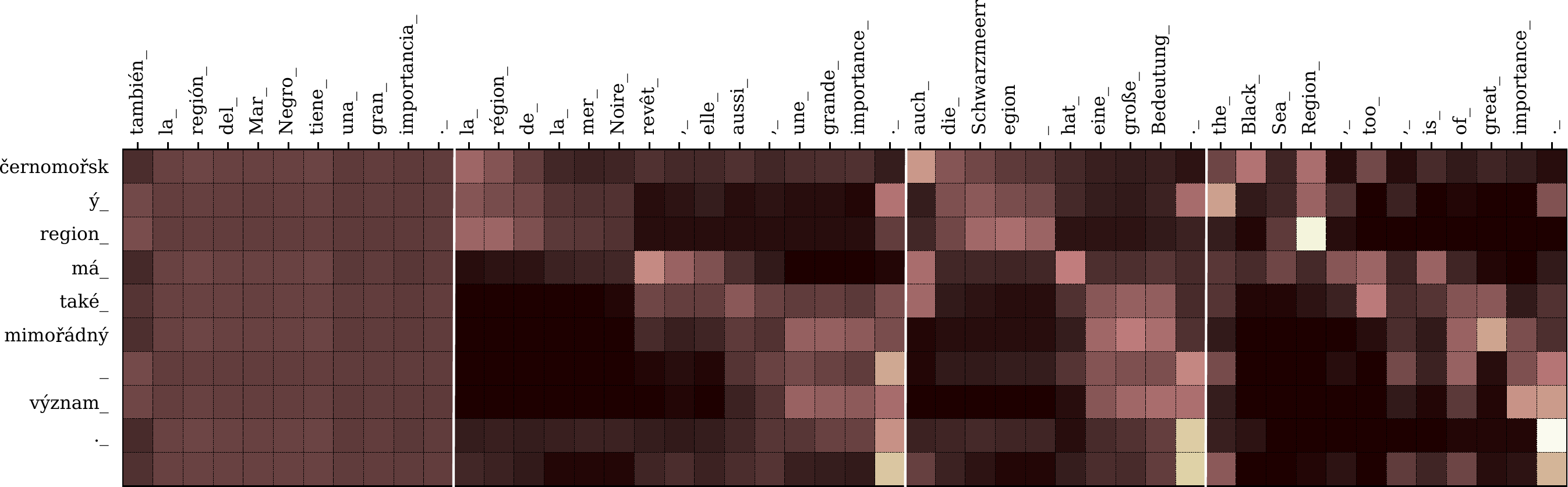} \\
b) parallel \\
\includegraphics[width=0.92\textwidth]{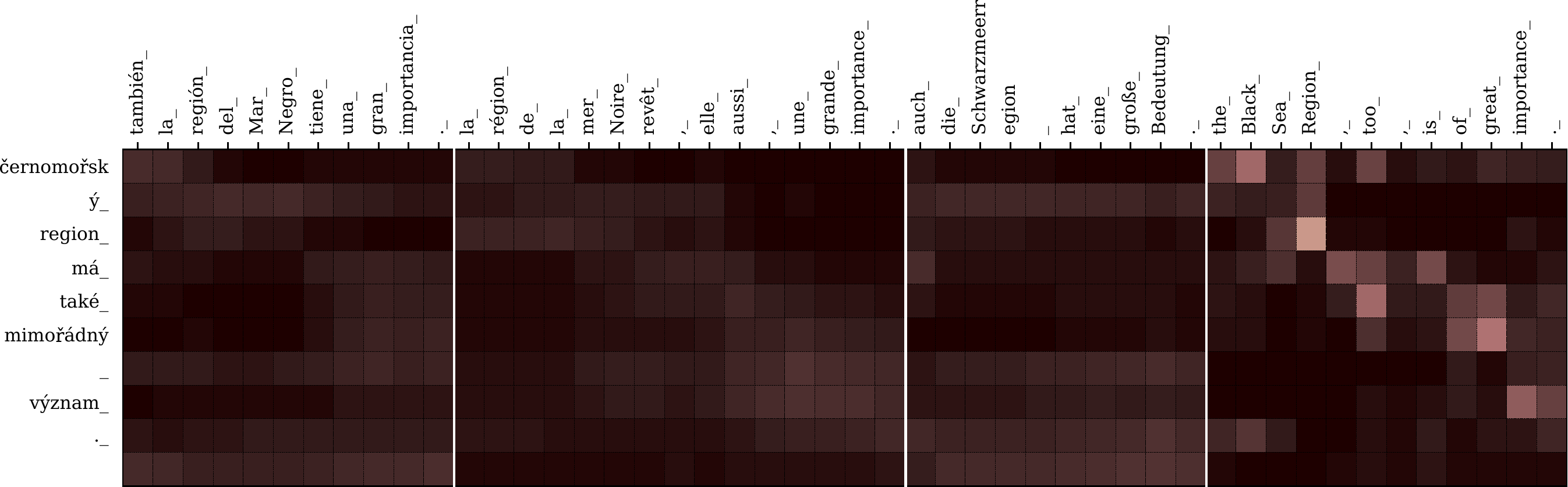} \\
c) flat \\
\includegraphics[width=0.92\textwidth]{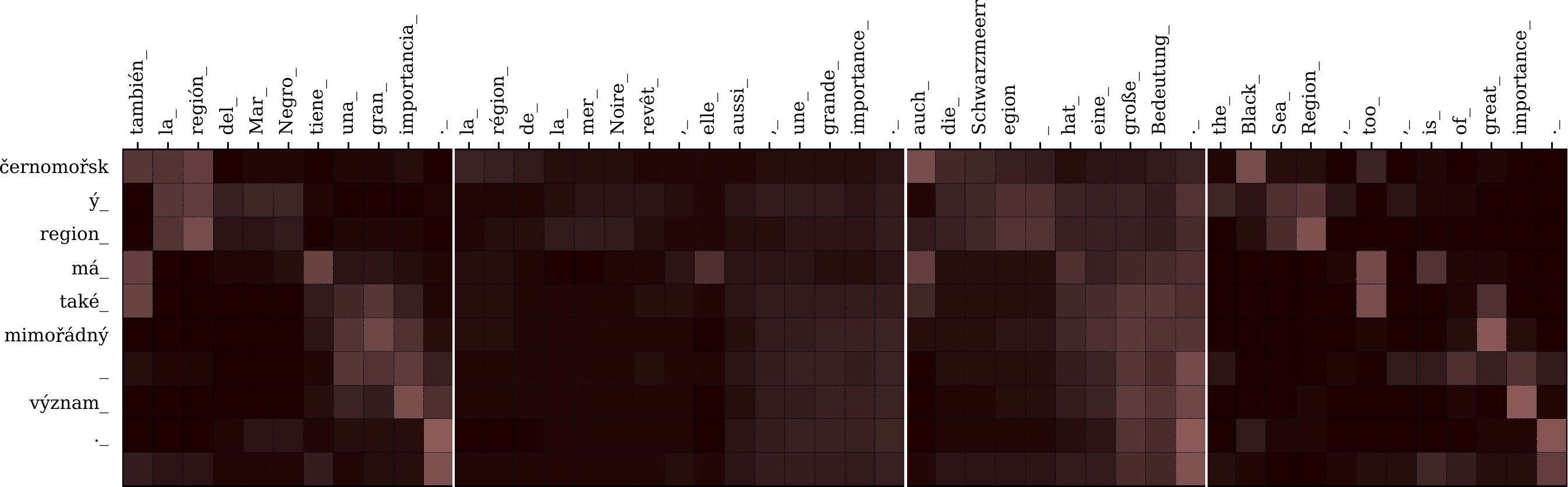} \\
d) hierarchical \\
\end{center}


\end{document}